\documentclass{article}

\usepackage[preprint]{neurips_2025}

\usepackage[utf8]{inputenc} 
\usepackage[T1]{fontenc}    
\usepackage{hyperref}       
\usepackage{url}            
\usepackage{booktabs}       
\usepackage{amsfonts}       
\usepackage{nicefrac}       
\usepackage{microtype}      
\usepackage[dvipsnames]{xcolor}         
\usepackage{algorithm}
\usepackage{algorithmic}
\usepackage{bm}
\usepackage{multirow}
\usepackage{multicol}
\usepackage{enumitem}
\usepackage{colortbl}
\usepackage{tabularx}
\usepackage{makecell}
\usepackage{algorithm}
\usepackage{algorithmic}

\usepackage{graphicx}
\usepackage{hhline}
\usepackage{pifont}
\usepackage{float}
\usepackage{amssymb,amsmath}
\usepackage{csquotes}
\usepackage{tcolorbox}
\tcbuselibrary{skins,breakable}

\newtcolorbox{framedquote}{
  enhanced,
  breakable,
  colback=gray!10,      
  colframe=black,     
  boxrule=1pt,        
  arc=2mm,            
  left=15pt, right=15pt, top=5pt, bottom=5pt,
  before skip=0em, after skip=1em
}

\newcommand{\down}[1]{$_{\color{Black}\downarrow #1}$}
\newcommand{\up}[1]{$_{\color{Black}\uparrow #1}$}
\definecolor{tableheadcolor}{RGB}{255,255,255}

\definecolor{mydarkblue}{rgb}{0,0.4,0.8} 
\definecolor{mylightblue}{rgb}{0.5,0.75,1}
\definecolor{NavyBlue}{HTML}{000080}
\hypersetup{
    colorlinks=true,
    linkcolor=red,
    citecolor=NavyBlue,
    filecolor=magenta,      
    urlcolor=magenta,
}

\usepackage{xspace}
\def\method{{\fontfamily{lmtt}\selectfont \textbf{MARCO}}\xspace}

\title{\method{}: Meta-Reflection with Cross-Referencing for Code Reasoning}

\author{%
  Yusheng Zhao\textsuperscript{1}, Xiao Luo\textsuperscript{2}, Weizhi Zhang\textsuperscript{3}, \\ \textbf{Zhiping Xiao\textsuperscript{4}, Wei Ju\textsuperscript{1}, Philip S. Yu\textsuperscript{3}, Ming Zhang\textsuperscript{1}} \\
  \textsuperscript{1} Peking University, \textsuperscript{2} University of California, Los Angeles,\\
  \textsuperscript{3} University of Illinois Chicago, \textsuperscript{4} University of Washington\\
  \texttt{yusheng.zhao@stu.pku.edu.cn}, 
\texttt{xiaoluo@cs.ucla.edu},
\texttt{patxiao@uw.edu},\\
\texttt{\{wzhan42,psyu\}@uic.edu}, 
\texttt{\{juwei,mzhang\_cs\}@pku.edu.cn} \\
}

\begin{document}

\maketitle

\begin{abstract}
The ability to reason is one of the most fundamental capabilities of large language models (LLMs), enabling a wide range of downstream tasks through sophisticated problem-solving. A critical aspect of this is code reasoning, which involves logical reasoning with formal languages (\emph{i.e.}, programming code). 
In this paper, we enhance this capability of LLMs by exploring the following question: \textbf{\textit{how can an LLM agent become progressively smarter in code reasoning with each solution it proposes, thereby achieving substantial cumulative improvement?}} 
Most existing research takes a static perspective, focusing on isolated problem-solving using frozen LLMs. In contrast, we adopt a cognitive-evolving perspective and propose a novel framework named \underline{M}et\underline{a}-\underline{R}eflection with \underline{C}r\underline{o}ss-Referencing (\method{}) that enables the LLM to evolve dynamically during inference through self-improvement.
From the perspective of human cognitive development, we leverage both \textbf{\textit{knowledge accumulation}} and \textbf{\textit{lesson sharing}}. In particular, to accumulate knowledge during problem-solving, we propose meta-reflection that reflects on the reasoning paths of the current problem to obtain knowledge and experience for future consideration. Moreover, to effectively utilize the lessons from other agents, we propose cross-referencing that incorporates the solution and feedback from other agents into the current problem-solving process.
We conduct experiments across various datasets in code reasoning, and the results demonstrate the effectiveness of \method{}.
\end{abstract}

\section{Introduction}
\begin{displayquote}
\textit{Go to bed a little smarter each day.}\hfill --- \textit{Warren Buffett}
\end{displayquote}
\label{sec:intro}
Large language models (LLMs) have achieved great success in understanding human instructions and generating texts \cite{achiam2023gpt, touvron2023llama, liu2024deepseek, guo2025deepseek}, enabling a plethora of downstream applications, including dialogue systems \cite{feng2023towards, chun2025llm}, automated summarization \cite{song2024learning, ghosh2024clipsyntel}, and code completion \cite{gu2023llm, fakhoury2024llm}. While they excel in retrieving and summarizing textual information, the ability to perform rational reasoning remains an important yet unfulfilled aspect when dealing with complex or logical tasks \cite{sun2023determlr, morishita2024enhancing}. 
Code reasoning \cite{zhao2025unveiling} is an important aspect of the reasoning ability of LLMs, involving the logical reasoning with formal languages (\emph{i.e.}, programming code) instead of natural languages. It requires abilities in both logical inference (rationality in content generation) and background knowledge (retrieval of past experiences) \cite{gu2024cruxeval, zhao2025unveiling}, making it a challenging task for LLMs.

\begin{figure}
    \centering
    \includegraphics[width=0.95\linewidth]{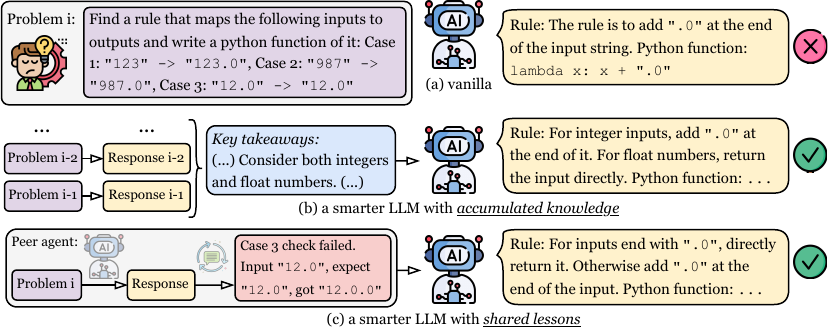}
    \vspace{-4mm}
    \caption{We adopt a cognitive-evolving  perspective and propose \method{} that enhances the code ability of an LLM through knowledge accumulation (b) and lesson sharing (c).}
    \vspace{-3mm}
    \label{fig:motivation}
\end{figure}

Recently, some research has investigated the LLMs' ability to solve complex or logical problems through reasoning. One line of research focuses on decomposing complex tasks into several manageable sub-tasks, using various data structures like chains \cite{wei2022chain, feng2023towards, zhang2024chain}, trees \cite{yao2023tree, cao2023probabilistic, ranaldi2024tree}, and graphs \cite{wen2023mindmap, lei2023boosting, besta2024graph, pandey2025adaptive}. Another line of research focuses on refining the solution iteratively through the feedback from various sources like reflection \cite{li2024selective, bașar2025well, li2025dora}, human guidance \cite{hu2023unlocking, dhole2023interactive}, or environmental input \cite{mower2024ros}, and adopts reinforcement learning for optimization \cite{zhang2024rest, meng2024llm, wang2024llm}. Despite their great success, existing research mainly adopts a \textbf{\textit{static perspective}}, focusing on problems themselves through task decomposition or solution refinement. However, the ability of the LLM itself during inference is not fully explored. Since human coders learn very fast through reading, understanding and writing the code, we aim to explore the following question:
\begin{quotation}
    \textit{How can an LLM agent become smarter at code reasoning with each response it generates, leading to substantial cumulative improvement?}
\end{quotation}
By answering this question, we adopt a \textbf{\textit{cognitive-evolving perspective}}, focusing on improving the ability of the LLM itself during problem-solving. To achieve this, we take both inter-problem knowledge accumulation and intra-problem lesson sharing, akin to human cognitive development.

As shown in Figure \ref{fig:motivation}, inter-problem knowledge accumulation requires the model to learn from the current problem-solving procedure to obtain transferable knowledge for future problems. To achieve this, we propose meta-reflection, which reflects on the reasoning process of the current problem and summarizes the experiences (\emph{e.g.} mistakes, common patterns) into concise takeaways. In future problem-solving, the condensed experiences are used as external knowledge to guide the thinking process.
Intra-problem lesson sharing enables the agent to learn from the lessons of other agents and use these experiences to guide the current problem-solving process. To achieve this, we propose cross-referencing that takes the solutions from other agents as well as the feedback into account when reasoning and proposing new solutions. We integrate the two components into a unified framework named \underline{M}et\underline{a}-\underline{R}eflection with \underline{C}r\underline{o}ss-Referencing (\method{}). Moreover, we perform extensive experiments across eight datasets and three sub-tasks in code reasoning (\emph{i.e.} induction, deduction and abduction), and the results validate the superiority of \method{} compared to baselines.

The contribution of this paper can be summarized as follows. \ding{172} \textbf{\textit{New Perspective}}: We propose a cognitive-evolving perspective that enhances the LLM’s capabilities through both inter- and intra-problem aspects, rather than relying on a static approach. \ding{173} \textbf{\textit{Novel Methodology}}: We propose \method{}, which consists of meta-reflection---reflecting on the current reasoning path for inter-problem knowledge accumulation---and cross-referencing, which incorporates the experiences of others into current problem-solving for intra-problem lesson sharing. \ding{174} \textbf{\textit{Extensive Experiments}}: We conduct extensive experiments across eight datasets and three sub-tasks in code reasoning, and the results demonstrate the effectiveness of the proposed \method{}.

\section{Preliminary}
\subsection{Problem Setup}
\label{sec:setup}
The execution of programs can be written as: $\mathcal I \xrightarrow{\mathcal F} \mathcal O$, forming a triplet of $\langle \mathcal I, \mathcal F, \mathcal O\rangle$, where $\mathcal I$ is the input, $\mathcal F$ is the function written in code, and $\mathcal O$ is the output of program execution. The goal of code reasoning is to infer one of the elements in the triplet (the target element) with the other two (source elements), forming three sub-tasks: induction, deduction, and abduction. Inductive code reasoning attempts to infer the function from the inputs and outputs, \emph{i.e.}, $\langle \mathcal I, \mathcal O\rangle\to \mathcal F$. Deductive code reasoning aims to deduce the output from the input and the function, \emph{i.e.}, $\langle \mathcal I, \mathcal F\rangle\to \mathcal O$. Abductive code reasoning aims to infer the input from the output and the function, \emph{i.e.}, $\langle \mathcal F, \mathcal O\rangle\to \mathcal I$. For simplicity, we denote the source elements of the problem as $\mathcal X$ and the target element as $\mathcal Y$. 

In practice, we are given a set of problems, \emph{i.e.}, $\{\mathcal X^i\}_{i=1}^N$, where $N$ is the number of problems. The large language model agent $\mathcal A_j$ ($j\in\{1,2,\cdots,M\}$) infers the solution $\mathcal Y^i_{j,T}$ with $\mathcal X^i$ through $T$ iterative reflection with feedback from the python interpreter. For each problem, we denote the corresponding answers as $\mathcal Y^i_{j,t}$, and the feedback as $\mathcal B^i_{j,t}$, where $t \in \{1,2,\cdots,T\}$. The reasoning path can be represented as $\mathcal P^i_{j,t} = (\mathcal X^i, \mathcal Y^i_{j,1}, \mathcal B^i_{j,1}, \mathcal Y^i_{j,2}, \mathcal B^i_{j,2}, \cdots, \mathcal Y^i_{j,t}, \mathcal B^i_{j,t})$. Under the static perspective, the model generates the next solution based on the current reasoning path of the current problem, \emph{i.e.}
\begin{equation}\label{eq:separate}
    \mathcal Y^i_{j,t}\sim\mathcal A_j(\mathcal Y \mid\mathcal P^i_{j,t-1}, \mathcal T_j),
\end{equation}
where $\mathcal A_j$ is the $j$-th LLM agent, and $\mathcal T_j$ is the textual prompt template of agent $\mathcal A_j$. Under this paradigm, the solutions $\mathcal Y^i_{j,T}$ are independent from each other in dimensions $i$ (inter-problem) and $j$ (intra-problem), and therefore, the model's ability is not improved when dealing with a sequence of problems $\{\mathcal X^i\}_{i=1}^N$. In comparison, we propose to incorporate both inter-problem and intra-problem information into the problem-solving process, \emph{i.e.} \vspace{-2mm}
\begin{equation}\label{eq:solution}
    \mathcal Y^i_{j,t}\sim\mathcal A_j(\mathcal Y \mid \mathcal P^i_{j,t-1}, \overbrace{\{\{\mathcal P^{i'}_{j,T}\}_{i'=1}^{i-1}\}_{j=1}^M}^{\text{inter-problem}},\overbrace{\{\mathcal P^i_{j',t-1}\}_{j'\in\{1,\cdots,M\} \setminus \{j\}}}^{\text{intra-problem}} , \mathcal T_j).
\end{equation}

\begin{figure}
    \centering
    \includegraphics[width=\linewidth]{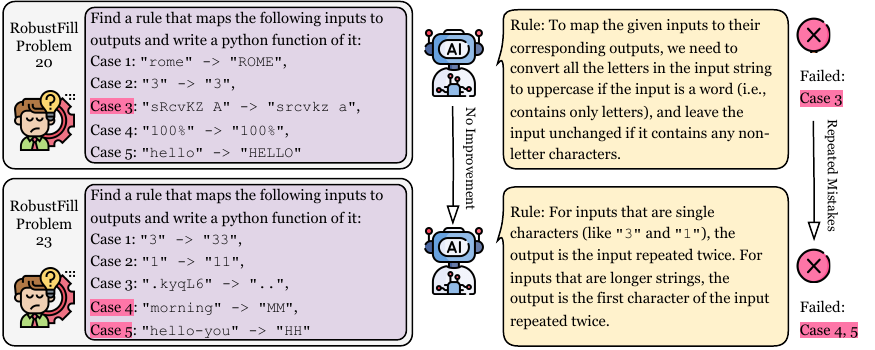}
    \vspace{-3mm}
    \caption{Existing methods adopt a static perspective, and the LLM agents do not improve during the problem-solving process, making repeated mistakes (in this case, lack of consideration of differences in upper/lower cases in constructing the transformation).}
    \vspace{-3mm}
    \label{fig:repeated}
\end{figure}

\subsection{Problems with Existing Methods: Repeated Mistakes}
Existing methods \cite{wei2022chain, li2023chain, zhao2025unveiling} take a static perspective, focusing on solving single problems independently. However, the ability of LLMs is not enhanced when solving a sequence of problems, resulting in repeated mistakes in the problem-solving process.

To illustrate this, we provide a concrete example of the problems with existing methods in Figure \ref{fig:repeated}. In problem 20, the LLM agent fails to generate the correct transformation due to a lack of consideration of the letter cases. Concretely, in strings with all lower-case letters, they are transformed to upper case, while in strings with both lower and upper cases, they are transformed to lower case. The generated transformation rule fails to consider the second case. A similar letter case mistake reappears in problem 23, where the cases of the letters are treated differently (for strings starting with lower case letters, the first characters are first raised to upper case and then repeated twice). However, as the problems are treated separately (as formulated in Eq. \ref{eq:separate}), the LLM agents cannot utilize previous experience for the current problem-solving process, making repeated mistakes. It is conceivable that if the agent is given guidance like "pay attention to different letter cases", it has a better chance of solving the problem correctly. Therefore, in this paper, we break the isolation from both inter-problem and intra-problem perspectives, incorporating the accumulated knowledge and shared lessons into the current problem-solving process (Eq. \ref{eq:solution} and its simplified form Eq. \ref{eq:final_sample}).

\section{Methodology}
\subsection{Framework Overview}
\begin{figure}
    \centering
    \includegraphics[width=\linewidth]{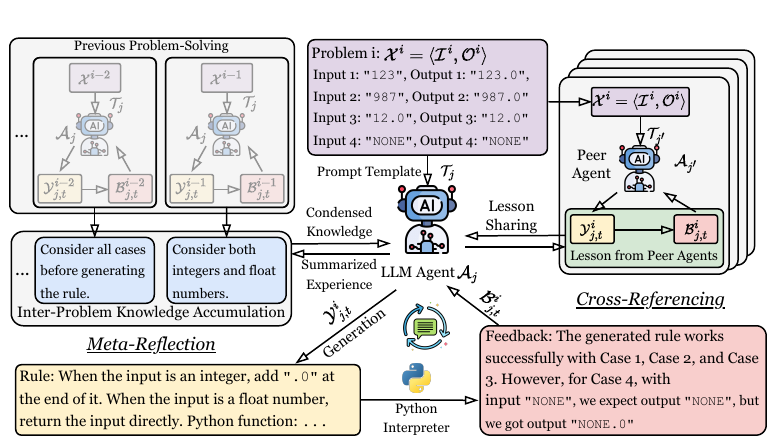}
    \vspace{-3mm}
    \caption{The overall framework of the proposed \method{}, which includes meta-reflection and cross-referencing. Meta-reflection summarizes previous problem-solving experiences into transferable knowledge accumulated for future usage. Cross-referencing enables the LLM agent to learn from the lessons of its peer agents so as to improve the current problem-solving process.}
    \vspace{-3mm}
    \label{fig:framework}
\end{figure}

Incorporating both inter- and intra-problem information into the current problem-solving process is appealing, but a naive solution would require excessively long contexts, leading to extremely high computation costs. Therefore, the \method{} framework addresses this challenge by adopting a cognitive-evolving perspective, using meta-reflection for inter-problem knowledge accumulation and cross-referencing for intra-problem lesson sharing, as shown in Figure \ref{fig:framework}. \method{} maintains a knowledge bank that stores summarized experiences of previous reasoning processes, with knowledge accumulating throughout the whole problem-solving process. When the LLM agents are given a specific problem, they first use the condensed knowledge from the knowledge bank to generate solutions. Subsequently, the code interpreter analyzes the solutions to provide feedback to the agents. For intra-problem lesson sharing, each agent is allowed to refer to the solutions and corresponding feedback of peer agents, thereby learning from the experiences of others.

\subsection{Meta-Reflection for Inter-Problem Knowledge Accumulation}
Previous efforts \cite{ji2023towards, bo2024reflective, zhaoevaluating} that utilize the reflection ability of LLMs primarily apply it within the context of a single problem. From a cognitive-evolving perspective, the capabilities of LLM agents do not improve as they solve a sequence of problems, which can lead to repeated mistakes (\emph{e.g.}, applying numerical operations to verbal strings). To address this, we propose meta-reflection that reflects on the reasoning paths to generate transferable knowledge for future reference.

Specifically, \method{} maintains a knowledge bank $\mathcal K$ which is initially empty ($\mathcal K^0=\varnothing$). When each agent finishes one problem (say, problem $\mathcal X^i$ and agent $\mathcal A_j$), it is provided with binary feedback $\mathcal B^i_j$ from the code interpreter about the preliminary checks of their solution, \emph{e.g.}, "All your answers are wrong for the given examples". Subsequently, the agent is asked to reflect on its reasoning process for problem $\mathcal X^i$ to generate a key takeaway that may be helpful for future problem-solving:
\begin{equation}\label{eq:summarize}
    \mathcal S_j^i=\operatorname{SUMMARIZE}(\mathcal A_j, \mathcal P^i_{j,T}, \mathcal B^i_j),
\end{equation}
where $\mathcal S^i_j$ is the summarized experience of this problem-solving experience. Then, $\mathcal S_j^i$ is added to the knowledge bank $\mathcal K^{i-1}$ to form $\mathcal K^i$, \emph{i.e.},
\begin{equation}\label{eq:append}
\mathcal K^i=\mathcal K^{i-1}\cup \{\mathcal S_{j}^{i}\}_{j=1}^M.
\end{equation}
A naive solution is to directly use this knowledge in the future. However, the knowledge bank grows during problem-solving, so incorporating all accumulated information becomes computationally burdensome. Additionally, key takeaways generated from different problem-solving instances may overlap, or be specific to a particular problem, rendering them non-transferable. Therefore, we employ a knowledge condenser that distills the knowledge bank into concise, transferable sentences,
\begin{equation}\label{eq:condense}
\hat{\mathcal K}^i = \operatorname{CONDENSE}(\mathcal A^{\textrm{condense}}, \mathcal K^i),
\end{equation}
where $\hat{\mathcal K}^i$ is the condensed knowledge and $\mathcal A^{\textrm{condense}}$ is the condenser agent. In practice, we condense the knowledge every $T_c$ problems to save computation and replace the previous $T_c$ problems with the condensed version to control the size of the knowledge bank. The condensed knowledge can then be used in future problem solving, simplifying Eq. \ref{eq:solution} as:
\begin{equation}\label{eq:meta-reflection}
    \mathcal Y^{i+1}_{j,t}\sim\mathcal A_j(\mathcal Y \mid \mathcal P^{i+1}_{j,t-1}, \hat{\mathcal K}^i,\{\mathcal P^{i+1}_{j',t-1}\}_{j'\in\{1,\cdots,M\} \setminus \{j\}} , \mathcal T_j).
\end{equation}

\subsection{Cross-Referencing for Intra-Problem Lesson Sharing}
Generating multiple solutions for each problem is a common strategy in recent research \cite{li2024enhancing, wan2024dynamic}. Nevertheless, the generation process is often isolated, and multiple agents can make similar mistakes in their reasoning paths. From the perspective of an LLM agent, when one of its peers makes a mistake, the agent should learn from that peer’s experience and avoid making similar mistakes in future iterations. As multiple agents are working on the same problem, the lessons learned from peers can be valuable for adjusting solutions.
Therefore, this paper proposes cross-referencing to enable intra-problem lesson sharing, allowing the agents to learn from the faults of others. 

Specifically, when agent $\mathcal A_j$ is given a problem $\mathcal X^i$, it falls into "proposal-feedback" iterations, where it first proposes a solution and receives feedback of it through the code interpreter. At iteration $t-1$, it receives feedback $\mathcal B_{j,t-1}^i$ of the proposal $\mathcal Y^i_{j,t-1}$. The proposal-feedback pair can be regarded as a "lesson", from which other agents may learn. To obtain a more compact version of this lesson for more efficient prompting, we extract the core part (\emph{e.g.}, python code) from the solution $\mathcal Y^i_{j,t-1}$ as $\hat{\mathcal Y}^i_{j,t-1}$, and the corresponding lesson can be denoted as $\langle\mathcal B_{j,t-1}^i, \hat{\mathcal Y}^i_{j,t-1}\rangle$.
When generating the $t$-th solution, the LLM agent incorporates the lessons shared by other agents, \emph{i.e.}, $\{\langle\mathcal B_{j,t-1}^i, \hat{\mathcal Y}^i_{j,t-1}\rangle\}_{j'\in\{1,\cdots,M\} \setminus \{j\}}$, and generates the solution with the simplified version of Eq. \ref{eq:meta-reflection} as follows:
\begin{equation}\label{eq:final_sample}
    \mathcal Y^{i}_{j,t}\sim\mathcal A_j(\mathcal Y \mid \mathcal P^{i}_{j,t-1}, \hat{\mathcal K}^i,\{\langle\mathcal B_{j,t-1}^i, \hat{\mathcal Y}^i_{j,t-1}\rangle\}_{j'\in\{1,\cdots,M\} \setminus \{j\}} , \mathcal T_j).
\end{equation}

\subsection{Summary}
The \method{} framework incorporates both meta-reflection and cross-referencing when dealing with a set of problems $\{\mathcal X^i\}_{i=1}^N$. For each problem $\mathcal X^i$, it first constructs an initial prompt using the problem and the condensed knowledge $\hat{\mathcal K}^{i-1}$. Subsequently, each agent is asked to generate solutions $\mathcal Y^i_{j,t}$, and the code interpreter is used to provide feedback $\mathcal B^i_{j,t}$. The agents then exchange their lessons $\langle\mathcal B_{j,t}^i, \hat{\mathcal Y}^i_{j,t}\rangle$, and adjust their solutions using information from their previous reasoning trajectories (\emph{i.e.}, chat history) and the lessons from peers. The iterative "proposal-feedback" process continues until a satisfactory solution is achieved (\emph{e.g.}, when the proposed function has passed all seen examples) or when reaching a certain computational budget (\emph{e.g.}, $T$ iterations).
\section{Experiments}
\subsection{Experimental Setup}
\label{sec:setup}
\textit{\textbf{Datasets.}} We conduct extensive experiments on eight datasets across three sub-tasks. For the inductive reasoning sub-task, we use four datasets: ListFunction \cite{rule2020child}, MiniARC \cite{kim2022playgrounds, qiu2023phenomenal}, RobustFill \cite{devlin2017robustfill}, and DeepCoder \cite{balog2016deepcoder}. Among these datasets, RobustFill and DeepCoder use specially designed domain-specific languages (DSLs, with details in Appendix \ref{sec:dsl}). For the deductive reasoning sub-task, we use two datasets: CRUXEval-O and LiveCodeBench-O that are built upon the CRUXEval \cite{gu2024cruxeval} and LiveCodeBench \cite{jain2024livecodebench} datasets by excluding the outputs. Similarly, for the abductive sub-task, we use CRUXEval-I and LiveCodeBench-I, built upon CRUXEval and LiveCodeBench respectively by excluding the inputs. 

\textit{\textbf{Experimental Details.}} We compare the proposed \method{} against three competitive reasoning methods: Chain-of-Code \cite{li2023chain}, Chain-of-Thought \cite{wei2022chain}, and RHDA \cite{zhao2025unveiling}. The comparison covers three representative LLM backbones, including ChatGPT-4o-mini, LLaMA 3 70B 
\cite{touvron2023llama}, Qwen 2.5 72B \cite{yang2024qwen2}. For the inductive sub-task, half of the input-output pairs are visible to the agent while the other half is not visible during the problem-solving process. For this sub-task, we report both accuracy (as measured by the ratio of correct input-output pairs and the total number of pairs) and problem accuracy (a problem is correctly solved if and only if all the input-output pairs of it are correct). 

\begin{table*}[t]
\centering
\resizebox{\textwidth}{!}{%
\def\arraystretch{1.1}
\begin{tabular}{l cc c cc c cc c cc}
\Xhline{1.2pt}
\rowcolor{tableheadcolor!20} & \multicolumn{2}{c}{ListFunction} && \multicolumn{2}{c}{MiniARC} &  & \multicolumn{2}{c}{RobustFill} && \multicolumn{2}{c}{DeepCoder} \\[-0.3pt]
\hhline{~--~--~--~--}\rowcolor{tableheadcolor!20}\multirow{-2}{*}{Models}
& Acc. & Prob. Acc. && Acc. & Prob. Acc. && Acc. & Prob. Acc. && Acc. & Prob. Acc. \\
\Xhline{1.2pt}
\textit{GPT 4o-mini} \\
\Xhline{1.0pt}
\rowcolor{gray!10} CoT & 36.70 & 26.40 && 3.85 & 2.31 && 37.39 & 17.39 && 16.32 & 7.29 \\
CoC & 34.50\down{2.20} & 26.40\up{0.00} && 1.79\down{2.06} & 0.77\down{1.54} && 50.43\up{13.04} & 21.74\up{4.35} && 18.75\up{2.43} & 8.33\up{1.04} \\
\rowcolor{gray!10} RHDA & 42.55\up{5.85} & 33.20\up{6.80} && 5.38\up{1.53} & 3.85\up{1.54} && 53.91\up{16.52} & 30.43\up{13.04} && 25.69\up{9.37} & 11.46\up{4.17} \\
\method{} (ours) & 49.35\up{12.65} & 39.60\up{13.20} && 5.90\up{2.05} & \underline{4.62}\up{2.31} && 57.39\up{20.00} & \underline{34.78}\up{17.39} && \textbf{35.07}\up{18.75} & \textbf{19.79}\up{12.50} \\
\Xhline{1.0pt}
\rowcolor{gray!10}\textit{Qwen 2.5 72B} & & & & & & & & & & & \\
\Xhline{1.0pt}
CoT & 39.50 & 30.00 && 5.64 & 3.85 && 36.52 & 21.74 && 18.06 & 7.29 \\
\rowcolor{gray!10}CoC & 14.40\down{25.10} & 10.00\down{20.00} && 1.28\down{4.36} & 0.00\down{3.85} && 20.87\down{15.65} & 8.70\down{13.04} && 18.40\up{0.34} & 8.33\up{1.04} \\
RHDA & 44.45\up{4.95} & 35.60\up{5.60} && 4.87\down{0.77} & 3.85\up{0.00} && 53.91\up{17.39} & 30.43\up{8.69} && 28.47\up{10.41} & 15.63\up{8.34} \\
\rowcolor{gray!10}\method{} (ours) & \textbf{54.90}\up{15.40} & \textbf{47.20}\up{17.20} && \underline{6.15}\up{0.51} & \underline{4.62}\up{0.77} && \textbf{63.48}\up{26.96} & \underline{34.78}\up{13.04} && 30.90\up{12.84} & \textbf{19.79}\up{12.50} \\
\Xhline{1.0pt}
\textit{LLaMA 3 70B} \\
\Xhline{1.0pt}
\rowcolor{gray!10}CoT & 35.20 & 25.20 && 4.62 & 3.08 && 35.65 & 17.39 && 14.93 & 9.38 \\
CoC & 34.85\down{0.35} & 26.80\up{1.60} && 0.00\down{4.62} & 0.00\down{3.08} && 31.30\down{4.35} & 13.04\down{4.35} && 19.44\up{4.51} & 7.29\down{2.09} \\
\rowcolor{gray!10}RHDA & 44.70\up{9.50} & 38.40\up{13.20} && 5.13\up{0.51} & 3.85\up{0.77} && 52.17\up{16.52} & 26.09\up{8.70} && 26.04\up{11.11} & \underline{16.67}\up{7.29} \\
\method{} (ours) & \underline{52.60}\up{17.40} & \underline{41.20}\up{16.00} && \textbf{8.46}\up{3.84} & \textbf{6.15}\up{3.07} && \underline{60.00}\up{24.35} & \textbf{39.13}\up{21.74} && \underline{32.29}\up{17.36} & \textbf{19.79}\up{10.41} \\
\Xhline{1.2pt}
\end{tabular}}
\vspace{-1mm}
\caption{Accuracies and problem accuracies on the inductive reasoning datasets, \emph{i.e.}, ListFunction, MiniARC, RobustFill, and DeepCoder. We \textbf{bold} the best results and \underline{underline} the second-best.}
\vspace{-3mm}
\label{tab:inductive}
\end{table*}

\subsection{Main Results}
We compare the proposed \method{} against baselines across three sub-tasks, \emph{i.e.}, inductive reasoning, deductive reasoning, and transductive reasoning. Results of inductive reasoning are shown in Table \ref{tab:inductive}, while the results of deductive and abductive reasoning are shown in Table \ref{tab:deductive_abductive}.

\textit{\textbf{Inductive Reasoning.}} Inductive reasoning requires the model to infer the mapping from the inputs to the outputs with the given examples of input-output pairs. In the experiments, we report both accuracy (how the proposed mapping fits the input-output pairs) and problem accuracy (how the problem is correctly solved with a mapping that fits all input-output pairs). According to the results in Table \ref{tab:inductive}, we can see that the proposed \method{} outperforms baselines in terms of both accuracy and problem accuracy by a large margin on various LLM backbones. This shows the overall effectiveness of the proposed \method{} on inductive reasoning. Additionally, we find that Chain-of-X methods (\emph{i.e.}, CoT and CoC) generally do not perform well on inductive reasoning. A possible explanation is that these methods focus on decomposing the problems, while the inductive code reasoning task requires a wholistic understanding of the input-output mapping. By comparison, reflection-based methods (\emph{i.e.}, RHDA and \method{}) are able to iteratively refine the solution with feedback, and this explains their better performance. The proposed \method{} accumulates knowledge through meta-reflection and learns from shared lessons through cross-referencing, leading to the best performance.

\textit{\textbf{Deductive Reasoning.}} Deductive code reasoning requires the model to deduce the output from the input and the program (in the form of a Python function). We show the performance of \method{} in deductive code reasoning compared to baselines in Table \ref{tab:deductive_abductive}. As can be seen from the results, our method significantly improves accuracy on deductive code reasoning, \emph{e.g.}, 28.3\% improvement relative to CoT in LiveCodeBench on LLaMA 3 70B. Additionally, we find that for models with weaker performance (\emph{e.g.}, LLaMA 3 70B), the proposed \method{} can achieve greater performance, unleashing the power of these models.

\begin{figure}[ht]
\vspace{-2mm}
  \begin{minipage}{0.36\textwidth}    
  \textit{\textbf{Abductive Reasoning.}} Abductive code reasoning requires the model to infer the inputs that lead to the corresponding outputs with the given program (function). In many cases, this would be harder than deductive code reasoning, and we can see from the results in Table \ref{tab:deductive_abductive} that the abductive reasoning accuracies are generally lower than deductive reasoning. Despite the difficulty, the proposed \method{} still achieves substantial improvement, \emph{e.g.}, 24.0\% improvement relative to CoT in CRUXEval on GPT4o-mini. Moreover, we find that the proposed \method{} achieves more improvement on abductive reasoning than inductive reasoning. One possible explanation is
  \end{minipage}%
  \hfill
  \begin{minipage}{0.62\textwidth}
    \centering
    \begin{table}[H]
\centering
\resizebox{\textwidth}{!}{%
\def\arraystretch{1.1}
\begin{tabular}{l cc c cc}
\Xhline{1.2pt}
\rowcolor{tableheadcolor!20} & \multicolumn{2}{c}{CRUXEval} && \multicolumn{2}{c}{LiveCodeBench} \\[-0.3pt]
\hhline{~--~--}\rowcolor{tableheadcolor!20}\multirow{-2}{*}{Models}
& Deductive & Abductive && Deductive & Abductive \\
\Xhline{1.2pt}
\textit{GPT 4o-mini} \\
\Xhline{1.0pt}
\rowcolor{gray!10} CoT & 80.50 & 65.12 && 79.41 & 50.00 \\
CoC & 81.00\up{0.50} & 64.88\down{0.24} && 80.39\up{0.98} & 50.00\up{0.00} \\
\rowcolor{gray!10} RHDA & 80.75\up{0.25} & 71.75\up{6.63} && 79.41\up{0.00} & 58.82\up{8.82} \\
\method{} (ours) &  \underline{83.62}\up{3.12} & \underline{80.75}\up{15.63} && \underline{87.25}\up{7.84} & \underline{64.71}\up{14.71} \\
\Xhline{1.0pt}
\rowcolor{gray!10}\textit{Qwen 2.5 72B} & & & & & \\
\Xhline{1.0pt}
CoT & 78.38 & 70.50 && 86.27 & 54.90 \\
\rowcolor{gray!10}CoC & 70.13\down{8.25} & 70.88\up{0.38} && 77.45\down{8.82} & 54.90\up{0.00} \\
RHDA & 80.38\up{2.00} & 74.63\up{4.13} && 78.43\down{7.84} & 58.82\up{3.92} \\
\rowcolor{gray!10}\method{} (ours) & \textbf{84.38}\up{6.00} & \textbf{81.75}\up{11.25} && \textbf{88.23}\up{1.96} & \textbf{65.69}\up{10.79} \\
\Xhline{1.0pt}
\textit{LLaMA 3 70B} \\
\Xhline{1.0pt}
\rowcolor{gray!10}CoT & 69.88 & 60.63 && 65.69 & 46.08 \\
CoC & 70.38\up{0.50} & 61.25\up{0.62} && 52.94\down{12.75} & 41.18\down{4.90} \\
\rowcolor{gray!10}RHDA & 78.25\up{8.37} & 71.75\up{11.12} && 66.67\up{0.98} & 55.88\up{9.80} \\
\method{} (ours) & 83.38\up{13.50} & 79.25\up{18.62} && 84.31\up{18.62} & \textbf{65.69}\up{19.61} \\
\Xhline{1.2pt}
\end{tabular}
} %
\caption{Prediction accuracies on the deductive and abductive reasoning. We \textbf{bold} the best and \underline{underline} the second-best.}
\label{tab:deductive_abductive}
\end{table}
  \end{minipage}
  \vspace{-4mm}
\end{figure}

that inferring the inputs from the outputs requires a deeper understanding of the functionality of the code, which calls for knowledge from past experiences and potential pitfalls. The proposed \method{} allows LLM agents to accumulate knowledge from past problem-solving processes through meta-reflection and to avoid potential pitfalls from shared lessons of peers through cross-referencing, leading to the best performance.

\subsection{Ablation Studies}

\begin{figure}[ht]
\vspace{-3mm}
  \begin{minipage}{0.43\textwidth}
  In this part, we investigate how the proposed meta-reflection and cross-referencing mechanisms affect the overall performance on code reasoning. Specifically, we adopt ListFunction for the inductive reasoning sub-task and LiveCodeBench for deductive and abductive reasoning sub-tasks. Three variants of \method{} are constructed: \method{}v1 removes meta-reflection from the original
  \end{minipage}%
  \hfill
  \begin{minipage}{0.55\textwidth}
    \centering
    \begin{table}[H]
\centering
\resizebox{\textwidth}{!}{%
\def\arraystretch{1.1}
\begin{tabular}{l cc c cc}
\Xhline{1.2pt}
\rowcolor{tableheadcolor!20} & \multicolumn{2}{c}{ListFunction} && \multicolumn{2}{c}{LiveCodeBench} \\[-0.3pt]
\hhline{~--~--}\rowcolor{tableheadcolor!20}\multirow{-2}{*}{Models}
& Acc. & Prob. Acc. && Deductive & Abductive \\
\Xhline{1.2pt}
\rowcolor{gray!10} \method{} & 49.35 & 39.60 && 87.25 & 64.71 \\
\method{}v1 & 46.60\down{2.75} & 36.80\down{2.80} && 84.31\down{2.94} & 62.75\down{1.96} \\
\rowcolor{gray!10} \method{}v2 & 42.90\down{6.45} & 32.80\down{6.80} && 86.27\down{0.98} & 62.75\down{1.96} \\
\method{}v3 & 41.30\down{8.05} & 32.80\down{6.80} && 81.37\down{5.88} & 61.76\down{2.95} \\
\Xhline{1.2pt}
\end{tabular}
} %
\caption{Ablation studies on the ListFunction and LiveCodeBench datasets across three reasoning sub-tasks.}
\label{tab:ablation}
\end{table}
  \end{minipage}
  \vspace{-5mm}
\end{figure}

version. \method{}v2 preserves the meta-reflection but excludes the knowledge condenser in Eq. \ref{eq:condense}. \method{}v3 disables cross-referencing from the original version. The results of the variants in comparison with the original version are shown in Table \ref{tab:ablation}. As can be seen from the results, meta-reflection, the condenser, and cross-referencing are all important for the overall performance of \method{}, as removing each of them leads to performance degradation.
Moreover, we find that the condense operation in Eq. \ref{eq:condense} is important for the success of meta-reflection. As reflecting on specific problem-solving may yield rules or experiences that only hold under certain conditions, the condense operation can identify the common knowledge and express it in a concise manner.

\begin{figure}[h]
    \centering
    \includegraphics[width=\linewidth]{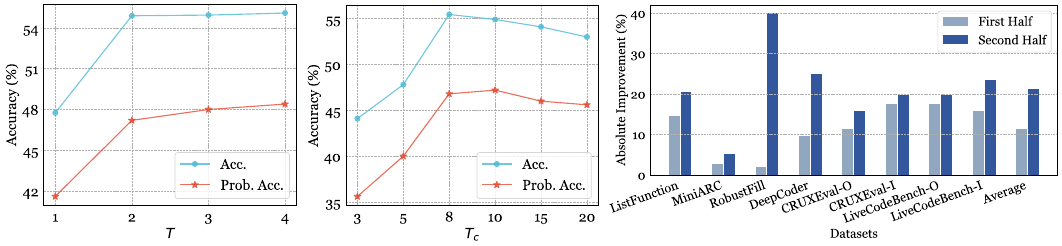}
    \vspace{-6mm}
    \caption{\textit{Left} and \textit{middle}: performance under different iterations and condensation periods in terms of accuracy and problem accuracy on the ListFunction dataset. \textit{Right}: the comparison of absolute improvements of \method{} and the baseline in both the first half and the second half of the datasets.}
    \vspace{-1mm}
    \label{fig:sensitivity}
\end{figure}

\subsection{Hyperparameter Analysis}
We then investigate the model's performance under different hyperparameters. Specifically, we focus on two hyperparameters: the number of iterations $T$, and the condensation period $T_c$, and the results on the ListFunction dataset using Qwen 2.5 72B are shown in Figure \ref{fig:sensitivity}, \textit{left}, and \textit{middle}. For the number of iterations, we find that the performance (\emph{i.e.}, accuracy and problem accuracy) plateaued after two iterations. This suggests that directly increasing the number of iterations and performing more reflections and revisions on the same problem is not sufficient to achieve better performance. Balancing the computation costs and performance, we set this hyperparameter to 2. As for the condensation period, we find that a moderate period of around 8 to 10 yields the best performance. For shorter periods, the condenser may not obtain enough knowledge to summarize, focusing on guidance hard to generalize. For longer periods, the model may fall short in updating the knowledge, and weaker adaptability causes a mild decrease in performance.

\subsection{Further Analysis}
\textit{\textbf{Effects of Meta-Reflection in the Problem-Solving Process.}} To better understand the effect of inter-problem knowledge accumulation in meta-reflection, we measure the absolute improvement (in accuracy) of \method{} compared to CoT in both the first half and the second half of the whole problem-solving process. The results using LLaMA 3 70B on various datasets are shown in Figure \ref{fig:sensitivity} (\textit{right}). As can be seen from the results, the absolute improvement of \method{} compared to the baseline increases during the whole problem-solving process, demonstrated by the fact that the second half experiences more improvement compared to the first half. These results suggest that, with the accumulated knowledge, the model is able to perform better with more understanding of the problems and additional guidance from previous reasoning processes. In other words, meta-reflection enhances the LLM's code reasoning ability during the problem-solving process.

\begin{figure}[t]
    \centering
    \includegraphics[width=\linewidth]{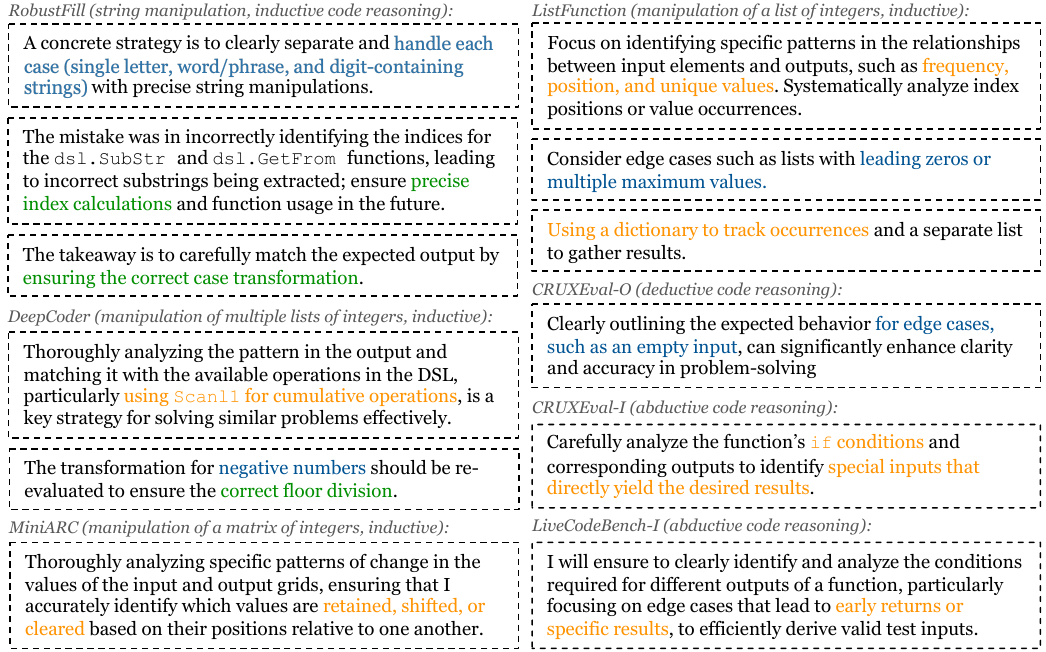}
    \vspace{-6mm}
    \caption{We present examples of the summarized reasoning experiences using meta-reflection on various datasets across three code reasoning sub-tasks (\emph{i.e.}, inductive, deductive, and abductive). The results suggest that meta-reflection can provide useful knowledge for future problem-solving.}
    \vspace{-4mm}
    \label{fig:meta-case}
\end{figure}

\textit{\textbf{Meta-Reflection Results Analysis.}} We also provide examples of meta-reflection results in Figure \ref{fig:meta-case}, from which we have the following observations. 
$\triangleright$ Firstly, by reflecting on the reasoning process, the LLM can generate guidance for handling different cases (marked as \textcolor[RGB]{66, 114, 163}{blue}). For example, in string manipulation (RobustFill, Example 1), the guidance can be \textit{"handle each case (single letter, word/phrase, and digit-containing strings) with precise string manipulations"}. Since separating letters and digits is common in string operations, this would be helpful for solving future problems. 
$\triangleright$ Secondly, meta-reflection allows the LLM to point out pitfalls that may recur in future problem-solving (marked as \textcolor[RGB]{62, 139, 38}{green}). For example, when dealing with integers (DeepCoder, Example 2), it points out \textit{"negative numbers should be re-evaluated to ensure the correct floor division"}. This is helpful for the model to avoid mistakes by paying more attention to certain operations (\emph{e.g.} floor division of negative numbers). 
$\triangleright$ Thirdly, meta-reflection helps find strategies that can be applied to future problem-solving (marked as \textcolor[RGB]{240, 152, 55}{orange}). For example, in abductive code reasoning, a simple strategy is to utilize special conditions as a shortcut to the return value. The LLM finds this simple strategy with meta-reflection, \emph{i.e.}, \textit{"carefully analyze the function's if conditions and corresponding outputs to identify special inputs that directly yield the desired results"} (CRUXEval-I).

\section{Related Works}
\textbf{LLM Reasoning.}
LLMs have achieved promising performance in understanding and generating texts. Recently, their ability to handle complex logical tasks through reasoning has drawn increasing attention. Inspired by the human thinking process, some studies attempt to break down complicated problems into more manageable sub-problems, which are handled one by one \cite{wei2022chain, xia2024beyond}. This line of research represents the reasoning process with various structures, including sequences \cite{wei2022chain, xu2025softcot}, trees \cite{yao2023tree, bi2024forest, ding2025dynamic}, and graphs \cite{wen2023mindmap, hsu2024thought, besta2024demystifying}. Another line of research focuses on searching for better solutions with LLMs according to the feedback provided by reflection \cite{renze2024self, xu2024jamplate}, consistency regularization \cite{wan2024dynamic}, human user \cite{laleh2024survey}, or physical environment \cite{luo2024integration, an2024iot}. To optimize the solutions, they often adopt reinforcement learning algorithms like Monte Carlo Tree Search \cite{zhang2024rest, gao2024interpretable, wang2024towards}, in combination with structures like trees \cite{hao2023reasoning} and graphs \cite{besta2024graph}. Compared to these works, which focus primarily on the problems via task decomposition or solution searching, we take a cognitive-evolving perspective that uses meta-reflection for inter-problem knowledge accumulation and cross-referencing for intra-problem lesson sharing to enhance the code ability of the LLM itself.

\textbf{LLM Reflection.}
The ability to perform reflection is an essential part of reasoning, involving evaluating the current solution and regenerating better answers \cite{renze2024self, kumar2024supporting}. In order to perform reflection, feedback is required either from LLMs themselves \cite{renze2024self, piche2024llms} or from the outside world (\emph{e.g.}, human user \cite{laleh2024survey}, simulated or physical environments \cite{an2024iot, rao2024collaborative}). While it may be attractive to reflect without external feedback, prior studies \cite{huang2023large, kumar2024training, zhao2025unveiling} suggest that, as analogous to the human learning process, self-correction without external feedback is hard to achieve without training and in some cases it may lead to performance degradation \cite{huang2023large}. In this work, we investigate the problem of code reasoning, which fortunately has compilers and Python interpreters as a source of external feedback \cite{jana2024cotran, izadi2024language}. Moreover, unlike previous methods \cite{shinn2023reflexion, zhaoevaluating, zhao2025unveiling} that use feedback solely for reflection on the current problem, this paper proposes meta-reflection that reflects on the whole reflection process to obtain transferable knowledge that can be used for future problem-solving and thus enhance the capabilities of the LLM from a cognitive-evolving perspective.

\textbf{LLM for Code.}
Recent years have witnessed increasing performance of LLMs on code understanding and generation \cite{wang2023codet5+, li2023large, liu2023your, du2024evaluating, chen2024survey, huang2024knowledge, jiang2024self}. These works are often designed for code generation based on human instructions or requirements \cite{du2024evaluating, jiang2024self, zhong2024can}, code completion based on existing code snippets \cite{guo2023longcoder, li2023cctest, izadi2024language}, or code debugging \cite{zhong2024debug, tian2024debugbench}. While this line of research has shown promising results, it largely relies on the LLMs' ability to recall \cite{wang2024unveiling} without much reasoning ability. By comparison, this work investigates a more challenging topic of code reasoning that requires both logical reasoning and recall capability in code-related tasks, involving induction, deduction, and abduction \cite{zhao2025unveiling}.

\section{Conclusion}
This paper investigates the problem of code reasoning and proposes \method{} that adopts a cognitive-evolving perspective, aiming to improve the code reasoning capabilities of LLMs during inference. Specifically, \method{} employs meta-reflection that reflects on the previous problem-solving processes to obtain transferable guidance for inter-problem knowledge accumulation. Moreover, it adopts cross-referencing that utilizes the failed experiences from peers to enable intra-problem lesson sharing. Extensive experiments on several benchmark datasets across three code reasoning sub-tasks demonstrate the effectiveness of the proposed \method{}.


\bibliographystyle{plain}
\bibliography{main.bib}

\begin{thebibliography}{10}

\bibitem{achiam2023gpt}
Josh Achiam, Steven Adler, Sandhini Agarwal, Lama Ahmad, Ilge Akkaya, Florencia~Leoni Aleman, Diogo Almeida, Janko Altenschmidt, Sam Altman, Shyamal Anadkat, et~al.
\newblock Gpt-4 technical report.
\newblock {\em arXiv preprint arXiv:2303.08774}, 2023.

\bibitem{an2024iot}
Tuo An, Yunjiao Zhou, Han Zou, and Jianfei Yang.
\newblock Iot-llm: Enhancing real-world iot task reasoning with large language models.
\newblock {\em arXiv preprint arXiv:2410.02429}, 2024.

\bibitem{balog2016deepcoder}
Matej Balog, Alexander~L Gaunt, Marc Brockschmidt, Sebastian Nowozin, and Daniel Tarlow.
\newblock Deepcoder: Learning to write programs.
\newblock {\em arXiv preprint arXiv:1611.01989}, 2016.

\bibitem{bașar2025well}
Erkan Bașar, Xin Sun, Iris Hendrickx, Jan de~Wit, Tibor Bosse, Gert-Jan De~Bruijn, Jos~A Bosch, and Emiel Krahmer.
\newblock How well can large language models reflect? a human evaluation of llm-generated reflections for motivational interviewing dialogues.
\newblock In {\em Proceedings of the 31st International Conference on Computational Linguistics}, pages 1964--1982, 2025.

\bibitem{besta2024graph}
Maciej Besta, Nils Blach, Ales Kubicek, Robert Gerstenberger, Michal Podstawski, Lukas Gianinazzi, Joanna Gajda, Tomasz Lehmann, Hubert Niewiadomski, Piotr Nyczyk, et~al.
\newblock Graph of thoughts: Solving elaborate problems with large language models.
\newblock In {\em Proceedings of the AAAI Conference on Artificial Intelligence}, volume~38, pages 17682--17690, 2024.

\bibitem{besta2024demystifying}
Maciej Besta, Florim Memedi, Zhenyu Zhang, Robert Gerstenberger, Guangyuan Piao, Nils Blach, Piotr Nyczyk, Marcin Copik, Grzegorz Kwa{\'s}niewski, J{\"u}rgen M{\"u}ller, et~al.
\newblock Demystifying chains, trees, and graphs of thoughts.
\newblock {\em arXiv preprint arXiv:2401.14295}, 2024.

\bibitem{bi2024forest}
Zhenni Bi, Kai Han, Chuanjian Liu, Yehui Tang, and Yunhe Wang.
\newblock Forest-of-thought: Scaling test-time compute for enhancing llm reasoning.
\newblock {\em arXiv preprint arXiv:2412.09078}, 2024.

\bibitem{bo2024reflective}
Xiaohe Bo, Zeyu Zhang, Quanyu Dai, Xueyang Feng, Lei Wang, Rui Li, Xu~Chen, and Ji-Rong Wen.
\newblock Reflective multi-agent collaboration based on large language models.
\newblock {\em Advances in Neural Information Processing Systems}, 37:138595--138631, 2024.

\bibitem{cao2023probabilistic}
Shulin Cao, Jiajie Zhang, Jiaxin Shi, Xin Lv, Zijun Yao, Qi~Tian, Juanzi Li, and Lei Hou.
\newblock Probabilistic tree-of-thought reasoning for answering knowledge-intensive complex questions.
\newblock {\em arXiv preprint arXiv:2311.13982}, 2023.

\bibitem{chen2024survey}
Liguo Chen, Qi~Guo, Hongrui Jia, Zhengran Zeng, Xin Wang, Yijiang Xu, Jian Wu, Yidong Wang, Qing Gao, Jindong Wang, et~al.
\newblock A survey on evaluating large language models in code generation tasks.
\newblock {\em arXiv preprint arXiv:2408.16498}, 2024.

\bibitem{chun2025llm}
Changwoo Chun, Daniel Rim, and Juhee Park.
\newblock Llm contextbridge: A hybrid approach for intent and dialogue understanding in ivsr.
\newblock In {\em Proceedings of the 31st International Conference on Computational Linguistics: Industry Track}, pages 794--806, 2025.

\bibitem{devlin2017robustfill}
Jacob Devlin, Jonathan Uesato, Surya Bhupatiraju, Rishabh Singh, Abdel-rahman Mohamed, and Pushmeet Kohli.
\newblock Robustfill: Neural program learning under noisy i/o.
\newblock In {\em International conference on machine learning}, pages 990--998. PMLR, 2017.

\bibitem{dhole2023interactive}
Kaustubh~D Dhole, Ramraj Chandradevan, and Eugene Agichtein.
\newblock An interactive query generation assistant using llm-based prompt modification and user feedback.
\newblock {\em arXiv preprint arXiv:2311.11226}, 2023.

\bibitem{ding2025dynamic}
Yifu Ding, Wentao Jiang, Shunyu Liu, Yongcheng Jing, Jinyang Guo, Yingjie Wang, Jing Zhang, Zengmao Wang, Ziwei Liu, Bo~Du, et~al.
\newblock Dynamic parallel tree search for efficient llm reasoning.
\newblock {\em arXiv preprint arXiv:2502.16235}, 2025.

\bibitem{du2024evaluating}
Xueying Du, Mingwei Liu, Kaixin Wang, Hanlin Wang, Junwei Liu, Yixuan Chen, Jiayi Feng, Chaofeng Sha, Xin Peng, and Yiling Lou.
\newblock Evaluating large language models in class-level code generation.
\newblock In {\em Proceedings of the IEEE/ACM 46th International Conference on Software Engineering}, pages 1--13, 2024.

\bibitem{fakhoury2024llm}
Sarah Fakhoury, Aaditya Naik, Georgios Sakkas, Saikat Chakraborty, and Shuvendu~K Lahiri.
\newblock Llm-based test-driven interactive code generation: User study and empirical evaluation.
\newblock {\em IEEE Transactions on Software Engineering}, 2024.

\bibitem{feng2023towards}
Yujie Feng, Zexin Lu, Bo~Liu, Liming Zhan, and Xiao-Ming Wu.
\newblock Towards llm-driven dialogue state tracking.
\newblock {\em arXiv preprint arXiv:2310.14970}, 2023.

\bibitem{gao2024interpretable}
Zitian Gao, Boye Niu, Xuzheng He, Haotian Xu, Hongzhang Liu, Aiwei Liu, Xuming Hu, and Lijie Wen.
\newblock Interpretable contrastive monte carlo tree search reasoning.
\newblock {\em arXiv preprint arXiv:2410.01707}, 2024.

\bibitem{ghosh2024clipsyntel}
Akash Ghosh, Arkadeep Acharya, Raghav Jain, Sriparna Saha, Aman Chadha, and Setu Sinha.
\newblock Clipsyntel: clip and llm synergy for multimodal question summarization in healthcare.
\newblock In {\em Proceedings of the AAAI Conference on Artificial Intelligence}, volume~38, pages 22031--22039, 2024.

\bibitem{gu2024cruxeval}
Alex Gu, Baptiste Rozi{\`e}re, Hugh Leather, Armando Solar-Lezama, Gabriel Synnaeve, and Sida~I Wang.
\newblock Cruxeval: A benchmark for code reasoning, understanding and execution.
\newblock {\em arXiv preprint arXiv:2401.03065}, 2024.

\bibitem{gu2023llm}
Qiuhan Gu.
\newblock Llm-based code generation method for golang compiler testing.
\newblock In {\em Proceedings of the 31st ACM Joint European Software Engineering Conference and Symposium on the Foundations of Software Engineering}, pages 2201--2203, 2023.

\bibitem{guo2023longcoder}
Daya Guo, Canwen Xu, Nan Duan, Jian Yin, and Julian McAuley.
\newblock Longcoder: A long-range pre-trained language model for code completion.
\newblock In {\em International Conference on Machine Learning}, pages 12098--12107. PMLR, 2023.

\bibitem{guo2025deepseek}
Daya Guo, Dejian Yang, Haowei Zhang, Junxiao Song, Ruoyu Zhang, Runxin Xu, Qihao Zhu, Shirong Ma, Peiyi Wang, Xiao Bi, et~al.
\newblock Deepseek-r1: Incentivizing reasoning capability in llms via reinforcement learning.
\newblock {\em arXiv preprint arXiv:2501.12948}, 2025.

\bibitem{hao2023reasoning}
Shibo Hao, Yi~Gu, Haodi Ma, Joshua~Jiahua Hong, Zhen Wang, Daisy~Zhe Wang, and Zhiting Hu.
\newblock Reasoning with language model is planning with world model.
\newblock {\em arXiv preprint arXiv:2305.14992}, 2023.

\bibitem{hsu2024thought}
Chi-Yang Hsu, Kyle Cox, Jiawei Xu, Zhen Tan, Tianhua Zhai, Mengzhou Hu, Dexter Pratt, Tianlong Chen, Ziniu Hu, and Ying Ding.
\newblock Thought graph: Generating thought process for biological reasoning.
\newblock In {\em Companion Proceedings of the ACM Web Conference 2024}, pages 537--540, 2024.

\bibitem{hu2023unlocking}
Zhiyuan Hu, Yue Feng, Anh~Tuan Luu, Bryan Hooi, and Aldo Lipani.
\newblock Unlocking the potential of user feedback: Leveraging large language model as user simulators to enhance dialogue system.
\newblock In {\em Proceedings of the 32nd ACM International Conference on Information and Knowledge Management}, pages 3953--3957, 2023.

\bibitem{huang2023large}
Jie Huang, Xinyun Chen, Swaroop Mishra, Huaixiu~Steven Zheng, Adams~Wei Yu, Xinying Song, and Denny Zhou.
\newblock Large language models cannot self-correct reasoning yet.
\newblock {\em arXiv preprint arXiv:2310.01798}, 2023.

\bibitem{huang2024knowledge}
Tao Huang, Zhihong Sun, Zhi Jin, Ge~Li, and Chen Lyu.
\newblock Knowledge-aware code generation with large language models.
\newblock In {\em Proceedings of the 32nd IEEE/ACM International Conference on Program Comprehension}, pages 52--63, 2024.

\bibitem{izadi2024language}
Maliheh Izadi, Jonathan Katzy, Tim Van~Dam, Marc Otten, Razvan~Mihai Popescu, and Arie Van~Deursen.
\newblock Language models for code completion: A practical evaluation.
\newblock In {\em Proceedings of the IEEE/ACM 46th International Conference on Software Engineering}, pages 1--13, 2024.

\bibitem{jain2024livecodebench}
Naman Jain, King Han, Alex Gu, Wen-Ding Li, Fanjia Yan, Tianjun Zhang, Sida Wang, Armando Solar-Lezama, Koushik Sen, and Ion Stoica.
\newblock Livecodebench: Holistic and contamination free evaluation of large language models for code.
\newblock {\em arXiv preprint arXiv:2403.07974}, 2024.

\bibitem{jana2024cotran}
Prithwish Jana, Piyush Jha, Haoyang Ju, Gautham Kishore, Aryan Mahajan, and Vijay Ganesh.
\newblock Cotran: An llm-based code translator using reinforcement learning with feedback from compiler and symbolic execution.
\newblock In {\em ECAI 2024}, pages 4011--4018. IOS Press, 2024.

\bibitem{ji2023towards}
Ziwei Ji, Tiezheng Yu, Yan Xu, Nayeon Lee, Etsuko Ishii, and Pascale Fung.
\newblock Towards mitigating hallucination in large language models via self-reflection.
\newblock {\em arXiv preprint arXiv:2310.06271}, 2023.

\bibitem{jiang2024self}
Xue Jiang, Yihong Dong, Lecheng Wang, Zheng Fang, Qiwei Shang, Ge~Li, Zhi Jin, and Wenpin Jiao.
\newblock Self-planning code generation with large language models.
\newblock {\em ACM Transactions on Software Engineering and Methodology}, 33(7):1--30, 2024.

\bibitem{kim2022playgrounds}
Subin Kim, Prin Phunyaphibarn, Donghyun Ahn, and Sundong Kim.
\newblock Playgrounds for abstraction and reasoning.
\newblock In {\em NeurIPS 2022 Workshop on Neuro Causal and Symbolic AI (nCSI)}, 2022.

\bibitem{kumar2024training}
Aviral Kumar, Vincent Zhuang, Rishabh Agarwal, Yi~Su, John~D Co-Reyes, Avi Singh, Kate Baumli, Shariq Iqbal, Colton Bishop, Rebecca Roelofs, et~al.
\newblock Training language models to self-correct via reinforcement learning.
\newblock {\em arXiv preprint arXiv:2409.12917}, 2024.

\bibitem{kumar2024supporting}
Harsh Kumar, Ruiwei Xiao, Benjamin Lawson, Ilya Musabirov, Jiakai Shi, Xinyuan Wang, Huayin Luo, Joseph~Jay Williams, Anna~N Rafferty, John Stamper, et~al.
\newblock Supporting self-reflection at scale with large language models: Insights from randomized field experiments in classrooms.
\newblock In {\em Proceedings of the eleventh ACM conference on learning@ scale}, pages 86--97, 2024.

\bibitem{laleh2024survey}
Alireza~Rashidi Laleh and Majid~Nili Ahmadabadi.
\newblock A survey on enhancing reinforcement learning in complex environments: Insights from human and llm feedback.
\newblock {\em arXiv preprint arXiv:2411.13410}, 2024.

\bibitem{lei2023boosting}
Bin Lei, Chunhua Liao, Caiwen Ding, et~al.
\newblock Boosting logical reasoning in large language models through a new framework: The graph of thought.
\newblock {\em arXiv preprint arXiv:2308.08614}, 2023.

\bibitem{li2023chain}
Chengshu Li, Jacky Liang, Andy Zeng, Xinyun Chen, Karol Hausman, Dorsa Sadigh, Sergey Levine, Li~Fei-Fei, Fei Xia, and Brian Ichter.
\newblock Chain of code: Reasoning with a language model-augmented code emulator.
\newblock {\em arXiv preprint arXiv:2312.04474}, 2023.

\bibitem{li2023large}
Jia Li, Chongyang Tao, Jia Li, Ge~Li, Zhi Jin, Huangzhao Zhang, Zheng Fang, and Fang Liu.
\newblock Large language model-aware in-context learning for code generation.
\newblock {\em ACM Transactions on Software Engineering and Methodology}, 2023.

\bibitem{li2025dora}
Kun Li, Tingzhang Zhao, Wei Zhou, and Songlin Hu.
\newblock Dora: Dynamic optimization prompt for continuous reflection of llm-based agent.
\newblock In {\em Proceedings of the 31st International Conference on Computational Linguistics}, pages 7546--7557, 2025.

\bibitem{li2024selective}
Ming Li, Lichang Chen, Jiuhai Chen, Shwai He, Jiuxiang Gu, and Tianyi Zhou.
\newblock Selective reflection-tuning: Student-selected data recycling for llm instruction-tuning.
\newblock In {\em Findings of the Association for Computational Linguistics ACL 2024}, pages 16189--16211, 2024.

\bibitem{li2024enhancing}
Weichen Li and Weimin Pan.
\newblock Enhancing chain-of-thought reasoning in large language models through text style diversity and prompt fusion.
\newblock In {\em Third International Conference on Electronic Information Engineering, Big Data, and Computer Technology (EIBDCT 2024)}, volume 13181, pages 226--232. SPIE, 2024.

\bibitem{li2023cctest}
Zongjie Li, Chaozheng Wang, Zhibo Liu, Haoxuan Wang, Dong Chen, Shuai Wang, and Cuiyun Gao.
\newblock Cctest: Testing and repairing code completion systems.
\newblock In {\em 2023 IEEE/ACM 45th International Conference on Software Engineering (ICSE)}, pages 1238--1250. IEEE, 2023.

\bibitem{liu2024deepseek}
Aixin Liu, Bei Feng, Bing Xue, Bingxuan Wang, Bochao Wu, Chengda Lu, Chenggang Zhao, Chengqi Deng, Chenyu Zhang, Chong Ruan, et~al.
\newblock Deepseek-v3 technical report.
\newblock {\em arXiv preprint arXiv:2412.19437}, 2024.

\bibitem{liu2023your}
Jiawei Liu, Chunqiu~Steven Xia, Yuyao Wang, and Lingming Zhang.
\newblock Is your code generated by chatgpt really correct? rigorous evaluation of large language models for code generation.
\newblock {\em Advances in Neural Information Processing Systems}, 36:21558--21572, 2023.

\bibitem{luo2024integration}
Xiaoyu Luo, Daping Liu, Fan Dang, and Hanjiang Luo.
\newblock Integration of llms and the physical world: Research and application.
\newblock In {\em Proceedings of the ACM Turing Award Celebration Conference-China 2024}, pages 1--5, 2024.

\bibitem{meng2024llm}
Silin Meng, Yiwei Wang, Cheng-Fu Yang, Nanyun Peng, and Kai-Wei Chang.
\newblock Llm-a*: Large language model enhanced incremental heuristic search on path planning.
\newblock {\em arXiv preprint arXiv:2407.02511}, 2024.

\bibitem{morishita2024enhancing}
Terufumi Morishita, Gaku Morio, Atsuki Yamaguchi, and Yasuhiro Sogawa.
\newblock Enhancing reasoning capabilities of llms via principled synthetic logic corpus.
\newblock {\em Advances in Neural Information Processing Systems}, 37:73572--73604, 2024.

\bibitem{mower2024ros}
Christopher~E Mower, Yuhui Wan, Hongzhan Yu, Antoine Grosnit, Jonas Gonzalez-Billandon, Matthieu Zimmer, Jinlong Wang, Xinyu Zhang, Yao Zhao, Anbang Zhai, et~al.
\newblock Ros-llm: A ros framework for embodied ai with task feedback and structured reasoning.
\newblock {\em arXiv preprint arXiv:2406.19741}, 2024.

\bibitem{pandey2025adaptive}
Tushar Pandey, Ara Ghukasyan, Oktay Goktas, and Santosh~Kumar Radha.
\newblock Adaptive graph of thoughts: Test-time adaptive reasoning unifying chain, tree, and graph structures.
\newblock {\em arXiv preprint arXiv:2502.05078}, 2025.

\bibitem{piche2024llms}
Alexandre Pich{\'e}, Aristides Milios, Dzmitry Bahdanau, and Chris Pal.
\newblock Llms can learn self-restraint through iterative self-reflection.
\newblock {\em arXiv preprint arXiv:2405.13022}, 2024.

\bibitem{qiu2023phenomenal}
Linlu Qiu, Liwei Jiang, Ximing Lu, Melanie Sclar, Valentina Pyatkin, Chandra Bhagavatula, Bailin Wang, Yoon Kim, Yejin Choi, Nouha Dziri, et~al.
\newblock Phenomenal yet puzzling: Testing inductive reasoning capabilities of language models with hypothesis refinement.
\newblock {\em arXiv preprint arXiv:2310.08559}, 2023.

\bibitem{ranaldi2024tree}
Leonardo Ranaldi, Giulia Pucci, Federico Ranaldi, Elena~Sofia Ruzzetti, and Fabio~Massimo Zanzotto.
\newblock A tree-of-thoughts to broaden multi-step reasoning across languages.
\newblock In {\em Findings of the Association for Computational Linguistics: NAACL 2024}, pages 1229--1241, 2024.

\bibitem{rao2024collaborative}
Sudha Rao, Weijia Xu, Michael Xu, Jorge Leandro, Ken Lobb, Gabriel DesGarennes, Chris Brockett, and Bill Dolan.
\newblock Collaborative quest completion with llm-driven non-player characters in minecraft.
\newblock {\em arXiv preprint arXiv:2407.03460}, 2024.

\bibitem{renze2024self}
Matthew Renze and Erhan Guven.
\newblock Self-reflection in llm agents: Effects on problem-solving performance.
\newblock {\em arXiv preprint arXiv:2405.06682}, 2024.

\bibitem{rule2020child}
Joshua~Stewart Rule.
\newblock {\em The child as hacker: building more human-like models of learning}.
\newblock PhD thesis, Massachusetts Institute of Technology, 2020.

\bibitem{shinn2023reflexion}
Noah Shinn, Federico Cassano, Ashwin Gopinath, Karthik Narasimhan, and Shunyu Yao.
\newblock Reflexion: Language agents with verbal reinforcement learning.
\newblock {\em Advances in Neural Information Processing Systems}, 36:8634--8652, 2023.

\bibitem{song2024learning}
Hwanjun Song, Taewon Yun, Yuho Lee, Jihwan Oh, Gihun Lee, Jason Cai, and Hang Su.
\newblock Learning to summarize from llm-generated feedback.
\newblock {\em arXiv preprint arXiv:2410.13116}, 2024.

\bibitem{sun2023determlr}
Hongda Sun, Weikai Xu, Wei Liu, Jian Luan, Bin Wang, Shuo Shang, Ji-Rong Wen, and Rui Yan.
\newblock Determlr: Augmenting llm-based logical reasoning from indeterminacy to determinacy.
\newblock {\em arXiv preprint arXiv:2310.18659}, 2023.

\bibitem{tian2024debugbench}
Runchu Tian, Yining Ye, Yujia Qin, Xin Cong, Yankai Lin, Yinxu Pan, Yesai Wu, Haotian Hui, Weichuan Liu, Zhiyuan Liu, et~al.
\newblock Debugbench: Evaluating debugging capability of large language models.
\newblock {\em arXiv preprint arXiv:2401.04621}, 2024.

\bibitem{touvron2023llama}
Hugo Touvron, Thibaut Lavril, Gautier Izacard, Xavier Martinet, Marie-Anne Lachaux, Timoth{\'e}e Lacroix, Baptiste Rozi{\`e}re, Naman Goyal, Eric Hambro, Faisal Azhar, et~al.
\newblock Llama: Open and efficient foundation language models.
\newblock {\em arXiv preprint arXiv:2302.13971}, 2023.

\bibitem{wan2024dynamic}
Guangya Wan, Yuqi Wu, Jie Chen, and Sheng Li.
\newblock Dynamic self-consistency: Leveraging reasoning paths for efficient llm sampling.
\newblock {\em arXiv preprint arXiv:2408.17017}, 2024.

\bibitem{wang2024llm}
Boyuan Wang, Yun Qu, Yuhang Jiang, Jianzhun Shao, Chang Liu, Wenming Yang, and Xiangyang Ji.
\newblock Llm-empowered state representation for reinforcement learning.
\newblock {\em arXiv preprint arXiv:2407.13237}, 2024.

\bibitem{wang2024towards}
Xiyao Wang, Linfeng Song, Ye~Tian, Dian Yu, Baolin Peng, Haitao Mi, Furong Huang, and Dong Yu.
\newblock Towards self-improvement of llms via mcts: Leveraging stepwise knowledge with curriculum preference learning.
\newblock {\em arXiv preprint arXiv:2410.06508}, 2024.

\bibitem{wang2024unveiling}
Yifei Wang, Yuheng Chen, Wanting Wen, Yu~Sheng, Linjing Li, and Daniel~Dajun Zeng.
\newblock Unveiling factual recall behaviors of large language models through knowledge neurons.
\newblock {\em arXiv preprint arXiv:2408.03247}, 2024.

\bibitem{wang2023codet5+}
Yue Wang, Hung Le, Akhilesh~Deepak Gotmare, Nghi~DQ Bui, Junnan Li, and Steven~CH Hoi.
\newblock Codet5+: Open code large language models for code understanding and generation.
\newblock {\em arXiv preprint arXiv:2305.07922}, 2023.

\bibitem{wei2022chain}
Jason Wei, Xuezhi Wang, Dale Schuurmans, Maarten Bosma, Fei Xia, Ed~Chi, Quoc~V Le, Denny Zhou, et~al.
\newblock Chain-of-thought prompting elicits reasoning in large language models.
\newblock {\em Advances in neural information processing systems}, 35:24824--24837, 2022.

\bibitem{wen2023mindmap}
Yilin Wen, Zifeng Wang, and Jimeng Sun.
\newblock Mindmap: Knowledge graph prompting sparks graph of thoughts in large language models.
\newblock {\em arXiv preprint arXiv:2308.09729}, 2023.

\bibitem{xia2024beyond}
Yu~Xia, Rui Wang, Xu~Liu, Mingyan Li, Tong Yu, Xiang Chen, Julian McAuley, and Shuai Li.
\newblock Beyond chain-of-thought: A survey of chain-of-x paradigms for llms.
\newblock {\em arXiv preprint arXiv:2404.15676}, 2024.

\bibitem{xu2024jamplate}
Xiaotong Xu, Jiayu Yin, Catherine Gu, Jenny Mar, Sydney Zhang, Jane~L E, and Steven~P Dow.
\newblock Jamplate: exploring llm-enhanced templates for idea reflection.
\newblock In {\em Proceedings of the 29th International Conference on Intelligent User Interfaces}, pages 907--921, 2024.

\bibitem{xu2025softcot}
Yige Xu, Xu~Guo, Zhiwei Zeng, and Chunyan Miao.
\newblock Softcot: Soft chain-of-thought for efficient reasoning with llms.
\newblock {\em arXiv preprint arXiv:2502.12134}, 2025.

\bibitem{yang2024qwen2}
An~Yang, Baosong Yang, Beichen Zhang, Binyuan Hui, Bo~Zheng, Bowen Yu, Chengyuan Li, Dayiheng Liu, Fei Huang, Haoran Wei, et~al.
\newblock Qwen2. 5 technical report.
\newblock {\em arXiv preprint arXiv:2412.15115}, 2024.

\bibitem{yao2023tree}
Shunyu Yao, Dian Yu, Jeffrey Zhao, Izhak Shafran, Tom Griffiths, Yuan Cao, and Karthik Narasimhan.
\newblock Tree of thoughts: Deliberate problem solving with large language models.
\newblock {\em Advances in neural information processing systems}, 36:11809--11822, 2023.

\bibitem{zhang2024rest}
Dan Zhang, Sining Zhoubian, Ziniu Hu, Yisong Yue, Yuxiao Dong, and Jie Tang.
\newblock Rest-mcts*: Llm self-training via process reward guided tree search.
\newblock {\em Advances in Neural Information Processing Systems}, 37:64735--64772, 2024.

\bibitem{zhang2024chain}
Xuan Zhang, Chao Du, Tianyu Pang, Qian Liu, Wei Gao, and Min Lin.
\newblock Chain of preference optimization: Improving chain-of-thought reasoning in llms.
\newblock {\em Advances in Neural Information Processing Systems}, 37:333--356, 2024.

\bibitem{zhaoevaluating}
Lili Zhao, Yang Wang, Qi~Liu, Mengyun Wang, Wei Chen, Zhichao Sheng, and Shijin Wang.
\newblock Evaluating large language models through role-guide and self-reflection: A comparative study.
\newblock In {\em The Thirteenth International Conference on Learning Representations}, 2025.

\bibitem{zhao2025unveiling}
Yuze Zhao, Tianyun Ji, Wenjun Feng, Zhenya Huang, Qi~Liu, Zhiding Liu, Yixiao Ma, Kai Zhang, and Enhong Chen.
\newblock Unveiling the magic of code reasoning through hypothesis decomposition and amendment.
\newblock {\em arXiv preprint arXiv:2502.13170}, 2025.

\bibitem{zhong2024can}
Li~Zhong and Zilong Wang.
\newblock Can llm replace stack overflow? a study on robustness and reliability of large language model code generation.
\newblock In {\em Proceedings of the AAAI Conference on Artificial Intelligence}, volume~38, pages 21841--21849, 2024.

\bibitem{zhong2024debug}
Li~Zhong, Zilong Wang, and Jingbo Shang.
\newblock Debug like a human: A large language model debugger via verifying runtime execution step-by-step.
\newblock {\em arXiv preprint arXiv:2402.16906}, 2024.

\end{thebibliography}

\end{document}